\documentclass[conference,a4paper]{IEEEtran}
\IEEEoverridecommandlockouts

\usepackage[hidelinks]{hyperref}
\usepackage[cmex10]{amsmath}
\usepackage{amssymb,amsfonts}
\usepackage{dblfloatfix}

\usepackage[ruled,vlined]{algorithm2e}
\usepackage{graphicx}
\graphicspath{{Figures/PDF/}{Figures/PNG/}}

\usepackage{animate}

\usepackage{booktabs}
\usepackage{siunitx}
\usepackage[numbers,compress]{natbib}
\usepackage{texnames}
\usepackage{bm,bbm}
\usepackage{orcidlink}
\usepackage{multirow}
\usepackage{subcaption} 

\begin{document}

\title{
Spatiotemporal Air Quality Mapping in Urban Areas Using Sparse Sensor Data, Satellite Imagery, Meteorological Factors, and Spatial Features
}


\author{	\IEEEauthorblockN{Osama Ahmad\orcidlink{0009-0003-2124-6114}, Zubair Khalid\orcidlink{0000-0001-7875-4687}, Muhammad Tahir\orcidlink{0000-0001-7162-568X}, and Momin Uppal\orcidlink{0000-0003-2122-1315}}
	\IEEEauthorblockA{\textit{School of Science and Engineering, Lahore University of Management Sciences},
		 Lahore 54792, Pakistan\\
	osama\_ahmad@lums.edu.pk, \,zubair.khalid@lums.edu.pk,\, tahir@lums.edu.pk,\, momin.uppal@lums.edu.pk}

}
\maketitle
\begin{abstract}
	Monitoring air pollution is crucial for protecting human health from exposure to harmful substances. Traditional methods of air quality monitoring, such as ground-based sensors and satellite-based remote sensing, face limitations due to high deployment costs, sparse sensor coverage, and environmental interferences. To address these challenges, this paper proposes a framework for high-resolution spatiotemporal Air Quality Index (AQI) mapping using sparse sensor data, satellite imagery, and various spatiotemporal factors. By leveraging Graph Neural Networks (GNNs), we estimate AQI values at unmonitored locations based on both spatial and temporal dependencies. The framework incorporates a wide range of environmental features, including meteorological data, road networks, points of interest (PoIs), population density, and urban green spaces, which enhance prediction accuracy. We illustrate the use of our approach through a case study in Lahore, Pakistan, where multi-resolution data is used to generate the air quality index map at a fine spatiotemporal scale. 
\end{abstract}

\begin{IEEEkeywords}
	Air pollution, recommendation system, semi-supervised, graph neural network.
\end{IEEEkeywords}

\section{Introduction}
According to the World Health Organization (WHO), air pollution is responsible for approximately seven million deaths annually, with urban areas being particularly vulnerable due to the high concentration of pollutants such as particulate matter (PM$_{2.5}$ and PM$_{10}$), sulfur dioxide (SO$_2$), nitrogen dioxide (NO$_2$), ozone (O$_3$), and carbon monoxide (CO). These pollutants have severe health consequences, including respiratory and cardiovascular diseases~\cite{manisalidis2020environmental}. To mitigate these risks, continuous and real-time monitoring of outdoor air quality is crucial for public health and policy-making.
In recent years, South Asia cities have been suffering from high pollution levels because of fossil fuel burning, industrial emissions, and crop burning~\cite{pippal2023understanding}. In this context, this work focuses on generating a spatiotemporal air quality map from a limited~(sparse) number of spatially distributed sensors in developing cities.

There are various methods to directly or indirectly measure outdoor air pollution such as remote sensing~\cite{shelestov2018air} and internet-of-things (IoT) based sensors~\cite{dhingra2019internet}. Satellite-based approaches, such as those utilizing Aerosol Optical Depth (AOD), provide broader coverage but suffer from issues related to cloud interference, surface reflectance, and atmospheric conditions~\cite{ranjan2021review}. Furthermore, the identification of optimal sensor placement remains a difficult task~\cite{munteanu2023situ}, and large-scale deployment of sensors often incurs considerable resource costs.

Recently, machine learning-based frameworks have been proposed to estimate air quality~\cite{feng2022air}.  Graph neural networks (GNNs) and gated recurrent units (GRUs) are commonly used to model spatiotemporal dependencies~\cite{su2023effective}. Most of the research had studied the correlation of air pollution with different features, such as meteorological parameters~\cite{xu2021influence}, point-of-interests, and road networks with real-time sensors using neural networks~\cite{zheng2013u},~\cite{hsieh2015inferring}. The Gaussian process (GP) regression has been used to identify the hot spot of pollutants in Lahore~\cite{ashraf2019spatial}. After analyzing the land use land cover patterns in~\cite{bharath2023understanding}, the author stated that urban expansion had degraded the air quality. Zhan et al. derived a statistical correlation of air pollution with natural and socioeconomic factors in China across different seasons~\cite{zhan2018driving}. Their studies claim that natural factors are more important and the location of air pollutant hot spots may vary from season to season. From the regression analysis, increasing wind speed, gross domestic product (GDP), and forest area in a region reduced air pollution~\cite{xu2019understanding}. To deploy new sensors in recommendation systems where AQI at unknown nodes is predicted using semi-supervised learning was employed to train the neural network~\cite{zheng2013u},~\cite{hsieh2015inferring},~\cite{yang2018sensor}. These methods studied those geographical factors that had a positive correlation with air pollution. In urban areas, green spaces and water bodies can be included to precisely predict AQI values. In an urban environment,~\cite{sorek2022deep}  used green spaces to determine their effect on air quality (AQ) estimation.  

Noting the recent development, this work proposes a framework for generating a high-resolution spatiotemporal AQI map for urban regions with sparse sensor data. By incorporating a variety of spatiotemporal factors—including meteorological data (at multiple resolutions), road networks, points of interest (PoIs), population density, and urban green spaces, our model captures the spatiotemporal dynamics of air quality across a region. Additionally, we integrate satellite-derived AOD data to enhance the ability of the model to infer air quality at unmonitored nodes. We evaluate the model on the Lahore region and provide insights into the impact of multi-resolution data and urban features on AQI prediction.

\begin{figure}[!t]
    \vspace{-2mm}
    \centering
    \includegraphics[width=0.95\linewidth]{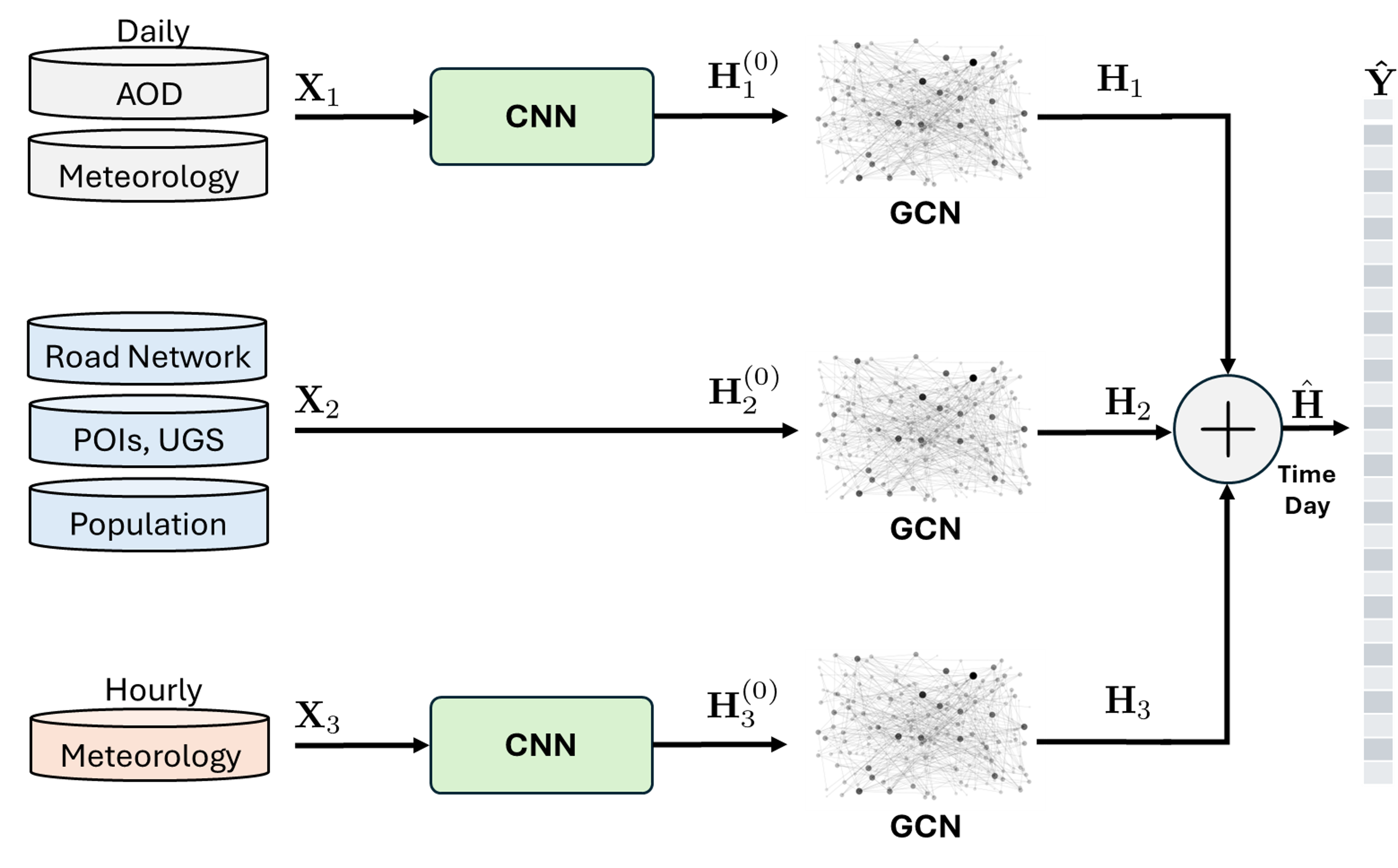}
    \caption{Block diagram of the proposed learning network. Temporal features pass through CNNs, producing \( \mathbf{H}^{(0)}_k \), while spatial features are processed using GCNs, producing \( \mathbf{H}_k \). }
    \vspace{-2mm}
    \label{fig:block_diagram}
\end{figure}

\section{Proposed Methodology}
\subsection{Problem Formulation}
This work focuses on estimating measurements at unlabeled nodes in a known static graph $\mathcal{G}$. Let $X$, $Y$, and $Y_L$ represent the features (inputs) to the graph, measurements at all nodes, and measurements at labeled nodes, respectively. The goal is to infer $Y_U$, the measurements at unlabeled nodes. The joint probability distribution for learning the model parameters is given by
\begin{equation}
    p(\Theta, Y_L, X) = p(\Theta \mid Y_L, X) \, p(Y_L \mid X) \, p(X),
\end{equation}
where $\Theta$ represents the model parameters. To estimate $Y_U$, we use the posterior distribution
 $   p(Y_U \mid X, \mathcal{G}, \Theta),$
which integrates the spatial and temporal dependencies of the data encoded in the graph structure $\mathcal{G}$.

\subsection{System Model}
We consider a graph $\mathcal{G} = (V, E, X)$, where $V$ is a set of nodes (grid cells in the region of interest) with a total number of nodes $|V| = N$, $E$ is a set of edges connecting the nodes, and $X$ represents the feature vectors associated with each node. 

The region of interest is divided into grid cells, each representing a node in the graph. The adjacency matrix $\mathbf{A}$ is constructed using a thresholded Gaussian kernel~\cite{shuman2013emerging} with elements expressed as
\begin{equation}
    a_{i,j} = 
    \begin{cases} 
    \exp\left(-\frac{d_{ij}^2}{\sigma^2}\right), & \text{if } \exp\left(-\frac{d_{ij}^2}{\sigma^2}\right) \ge r, \\
    0, & \text{otherwise},
    \end{cases}
\end{equation}
where $d_{ij}$ denotes the Euclidean distance between nodes $i$ and $j$, $\sigma$ controls the scale of the kernel, and $r$ is the threshold for adjacency. The degree of the weighted adjacency matrix is formulated as 
 $\mathbf{D}_{ii}=\sum_j  \mathbf{A}_{(i,j)}$, which is used to define the normalized adjacency matrix~\cite{kipf2016semi} as
 $\mathbf{\tilde{A}}=\mathbf{D}^{-\frac{1}{2}}(\mathbf{A+I_N})\mathbf{D}^{-\frac{1}{2}}$. 
 
Each node in the graph is associated with spatio-temporal features such as temperature, pressure, humidity, and additional spatial characteristics like population density, road networks, and green spaces. These features are encoded as inputs to the graph to model complex dependencies across nodes. The features are categorized into types such as spatial and spatiotemporal. Spatial features have only spatial dependencies while spatiotemporal have both time and space.

\subsection{Learning Framework}
The overall learning network is shown in Fig.~\ref{fig:block_diagram}. To model spatial (or topological) features and capture temporal dynamics, we incorporate Graph Convolutional Networks (GCNs) and Convolutional Neural Networks (CNNs), respectively. The spatiotemporal data is characterized by two different resolutions. The feature matrix $\mathbf{X}_1 \in \mathbb{R}^{C_1 \times T_1 \times N}$ consists of aerosol optical depth (AOD), land surface temperature (LST), surface pressure, humidity, wind speed, and wind direction, all with a one-day resolution. In contrast, $\mathbf{X}_3 \in \mathbb{R}^{C_3 \times T_3 \times N}$ has a one-hour resolution, where its features include temperature, humidity, and pressure. Here, $C_1$ or $C_3$ represents the total number of features and $T_1$ or $T_3$ is the number of time-points (days or hours) of the historical data respectively.

The processing of the spatiotemporal data in the network is given by
\begin{equation}
    \mathbf{H}^{(l+1)}_k = f \left(\mathbf{\tilde{A}} \mathbf{H}^{(l)}_k \mathbf{W}^{(l)}_k \right), \quad \mathbf{H}^{(0)}_k = f \left(\text{conv}(\mathbf{X}_k, \mathcal{W}_k)\right),
    \label{Eq:processing_GCN}
\end{equation}
where $k \in \{1,3\}$, and $\mathbf{H}^{(0)}_k$ denotes the initial input data for each feature set $\mathbf{X}_k$. In  \eqref{Eq:processing_GCN}, $\mathbf{W}_k$ and $\mathcal{W}_k$ represent the weights of the GCN and CNN for feature $\mathbf{X}_k$, respectively, $f$ denotes the activation function and $l$ is the layer of the GCN.

Spatial data is processed using GCN layers as
\begin{equation}
    \mathbf{H}^{(l+1)}_2 = f \left( \mathbf{\tilde{A}} \mathbf{H}^{(l)}_2 \mathbf{W}^{(l)}_2 \right),
\end{equation}
with the initial feature $\mathbf{H}_2^{(0)} = \mathbf{X}_2$. We combine the output of GCNs as
\begin{equation}
    \mathbf{\hat{H}} = \alpha \mathbf{H}_1 + \beta \mathbf{H}_2 + \gamma \mathbf{H}_3,
\end{equation}
which is then vectorized before concatenating with the time of the day and the day of the week and then passing through a GCN layer to obtain the predicted output $\mathbf{\hat{Y}}$ over $N$ number of nodes. Here $\alpha$, $\beta$, and $\gamma$ are learnable parameters. The ground truth $\mathbf{Y}$ represents the real-time AQI values obtained from the sensors at a given time. To train the network, the following total loss function is minimized~\cite{feng2020graph}
\begin{equation}
    \mathcal{L}_{\rm{total}} = \mathcal{L}_{\rm{sup}} + \lambda \mathcal{L}_{\rm{reg}},
\end{equation}
where $\mathcal{L}_{\rm{sup}}$ is the supervised loss computed as the mean squared error (MSE) on the labeled nodes $V_L$ and $\lambda$ is the penalty term for the unsupervised~(smoothness) loss given by~\cite{bontonou2019introducing} 
\begin{equation*}
    \mathcal{L}_{\rm{reg}}=\sum_{i,j \in V} a_{ij} (\hat{Y}_i-\hat{Y}_j)^2. 
\end{equation*}

\section{Implementation}
\begin{figure}[!t]
    \vspace{-2mm}
    \centering
    \includegraphics[width=0.9\linewidth]{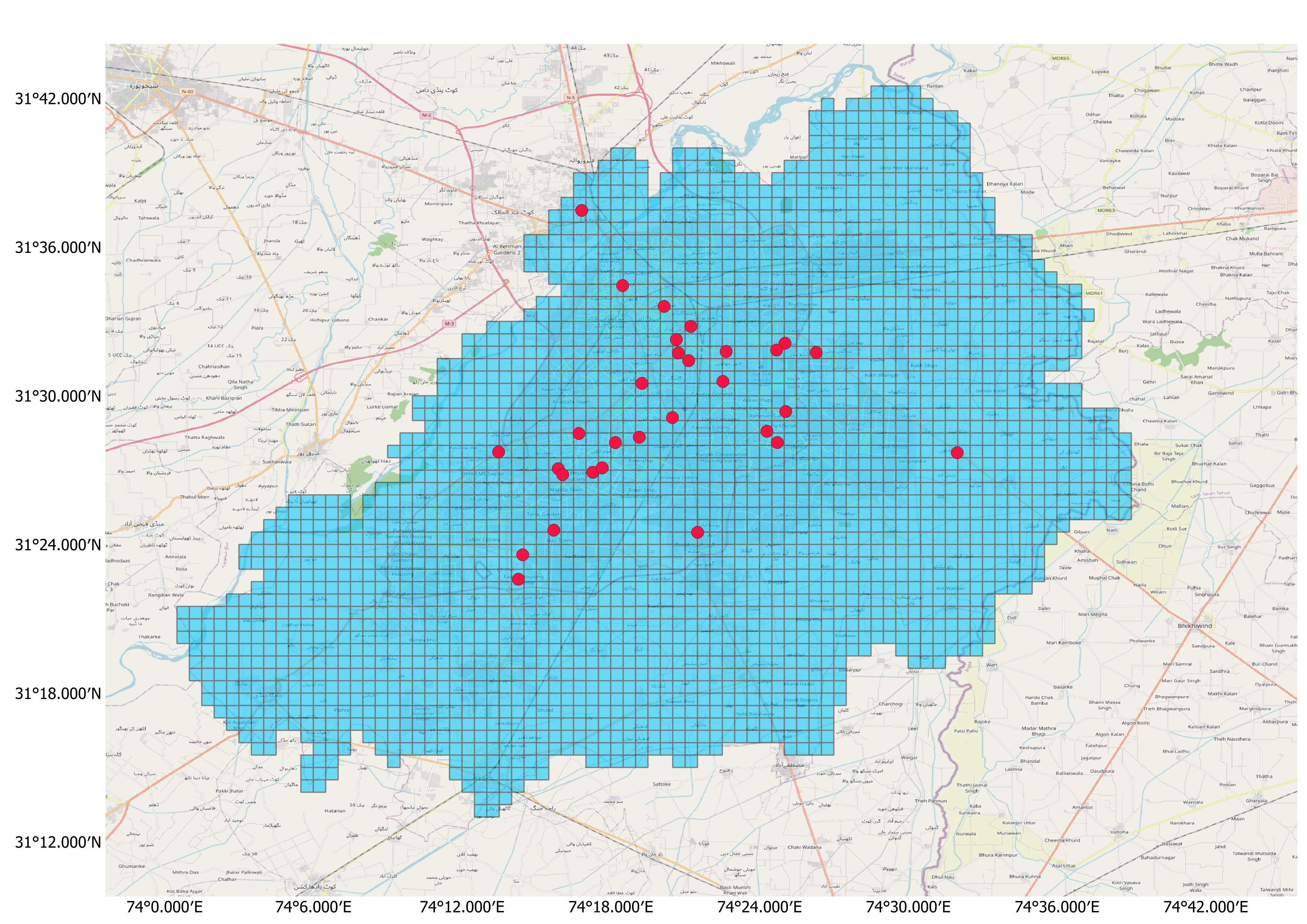}
    \caption{Distribution of 30 AQI Sensors in Lahore.}
    \vspace{-2mm}
    \label{fig:sensor_loc}
\end{figure}

We consider Lahore District, Pakistan's second-most populous city, covering 1,772 km² at 31.5497°N, 74.3436°E, as a study area. We divide the Lahore region into $2401$ grid cells~(each with a spatial resolution of 1km$\times$1km), where each grid cell is labeled from 1 to 2401. In Lahore, $30$ monitoring stations are measuring real-time AQI, covering only $1.25\%$ spatial area in our grid division. The spatial distribution of sensors in the Lahore region is shown in Fig.~\ref{fig:sensor_loc}. 

\subsection{Data Acquisition}
Data of various types, including air quality, meteorological conditions, road networks, points of interest, population density, and urban green spaces, are collected from multiple sources. These include geo-satellites (MODIS and ERA5), web services (Google Maps and OpenStreetMap), real-time sensors, and open-source datasets.
 
\subsubsection{Particulate Matter Data} We manually log the 2.5$\mu m$ particulate matter (PM$_{2.5}$) and AQI data from IQAir at hourly rate resolution from $20/12/2024$ to $31/12/2024$.  Aerosol optical depth (AOD) is extracted from MODIS (Moderate Resolution Imaging Spectroradiometer) every day with a spatial resolution of $1\;km$. It contains the aerosol size of $0.44 \;\mu m$ and $0.55\;\mu m$. We employ the data of the MCD19A2 product that utilizes the multi-angle implementation of atmospheric correction (MAIAC) algorithm for cloud masking and corrects atmospheric effects over both dark vegetated surfaces and bright deserts.
\subsubsection{Meteorological Data} Atmospheric data has been extracted from ERA5~\cite{hersbach2018era5} with a spatial resolution of $9 \;km$ with the daily interval.  We concatenated AOD, land surface temperature, pressure, relative humidity, wind speed, and wind direction to capture complex patterns. IQAir sensors provide temperature, pressure, and humidity at an interval of one hour. We use the inverse distance weighting to interpolate the data at the nodes where the measurements are not available.
\begin{table}[!b]
    \centering
    \caption{Types of PoIs}
    \label{tab:pois_list}
    \begin{tabular}{|c|c|}
    \hline
         P1: Factories & P5: Banks and financial institutions  \\
         \hline
         P2: Fuel station &P6: Supermarkets and shopping centers \\
         \hline
         P3: Utility stores & P7: Health centers and pharmacies\\
         \hline
         P4: Education   &P8: Food Points\\
         \hline
    \end{tabular}
\end{table}
\subsubsection{Road Networks} Since the traffic congestion directly impacts air quality, we incorporate the road network to determine its correlation with the concentration of air pollutants. The aggregate length of smaller roads within the grid cell is used as a node feature. The total number of T-intersections and multi-leg intersections within a region are also included in the feature matrix. 
\subsubsection{Point-of-Interests (POIs)} People and traffic congestion depend on the infrastructure and point-of-interests (PoIs). Such PoIs determine the congestion patterns during rush hours in our daily routine. TABLE~\ref{tab:pois_list} presents distinct eight PoIs, extracted from the Google maps and incorporated in our work. 
\subsubsection{Population Count} In urban areas, densely populated areas have a positive correlation with polluting agents. We collected an estimated high-resolution population count provided by Landscan~\cite{landscan} with 30-arc seconds resolution. 
\subsubsection{Urban Green Spaces (UGS)} 
Urban green spaces (UGS) serve as natural filters, absorbing pollutants from the ambient environment and exhibiting a strong inverse correlation with air pollution. In this study, UGS data is incorporated into the model by summing the green areas within each grid cell. These spaces include various categories such as amenities, barriers, boundaries, land use, leisure areas, natural reserves, tourism sites, and parks, all extracted from OpenStreetMap (OSM).

We set $r = 0.01$ to construct the adjacency matrix and normalize the features using $(x - \mu) / \sigma$, where $\mu$ and $\sigma$ represent the mean and standard deviation of each feature, respectively. The model is implemented using a deep neural network framework in the PyTorch environment, with the Adam optimizer employed for training over 100 epochs. To ensure robustness, we perform two independent experimental runs. In each run, two sensors are randomly selected for testing, while the remaining sensors are used for training the model. In our experimentation, we use $T_1$ as 4-day historical data and $T_3$ as 24-hour historical data. We compute the mean absolute error (MAE), root mean square error (RMSE), mean absolute percentage error (MAPE) to evaluate the performance.

\begin{table*}[!t]
    \vspace{-2mm}
    \centering
    \caption{Performance evaluation metrics MAE, RMSE, and MAPE for different values of learning rate ($lr$) and $\lambda$. }
    \label{tab:parameter}
    \begin{tabular}{cc|ccc|ccc|ccc}
    \hline
    \multirow{2}{*}{$lr$}&\multirow{2}{*}{$\lambda$}& \multicolumn{3}{c}{Run 1}&\multicolumn{3}{c}{Run 2}&\multicolumn{3}{c}{Average}\\
     \cmidrule(r){3-11}
    &     & MAE&RMSE&MAPE&MAE&RMSE&MAPE&MAE&RMSE&MAPE   \\
         \hline
       $10^{-4}$&$10^{-4}$& $86.57$&$90.15$&$28.03\%$&$88.89$&$91.54$&$46.50\%$&$76.67$&$83.54$&$28.32\%$\\
       $10^{-4}$&$10^{-5}$&$86.58$&$90.17$&$28.03\%$&$57.38$&$57.22$&$25.95\%$&$75.00$&$81.97$&$27.61\%$\\
       $10^{-4}$  &$10^{-6}$&$84.93$&$88.73$&$27.74\%$&$55.38$&$57.23$&$25.95\%$&$73.23$&$78.30$&$28.54\%$\\
       $10^{-3}$  &$10^{-5}$&$88.74$&$93.11$&$28.53\%$&$53.01$&$54.75$&$24.33\%$&$79.18$&$85.31$&$25.82\%$\\
       $10^{-5}$  &$10^{-5}$&$85.33$&$90.22$&$26.95\%$&$88.74$&$91.42$&$46.18\%$&$74.18$&$80.68$&$31.77\%$\\
         \hline
    \end{tabular}
    \vspace{-4mm}
\end{table*}

\section{Results}

We evaluate the performance of the proposed framework under various combinations of learning rate ($lr$) and $\lambda$. For each experimental run, a random pair of sensors (out of 30) is selected for validation, while the sensor sequence remains fixed to isolate the effects of hyperparameters.

\begin{figure}[!t]
    \centering
    \vspace{-2mm}
    \animategraphics[loop, autoplay, scale=0.57, controls]{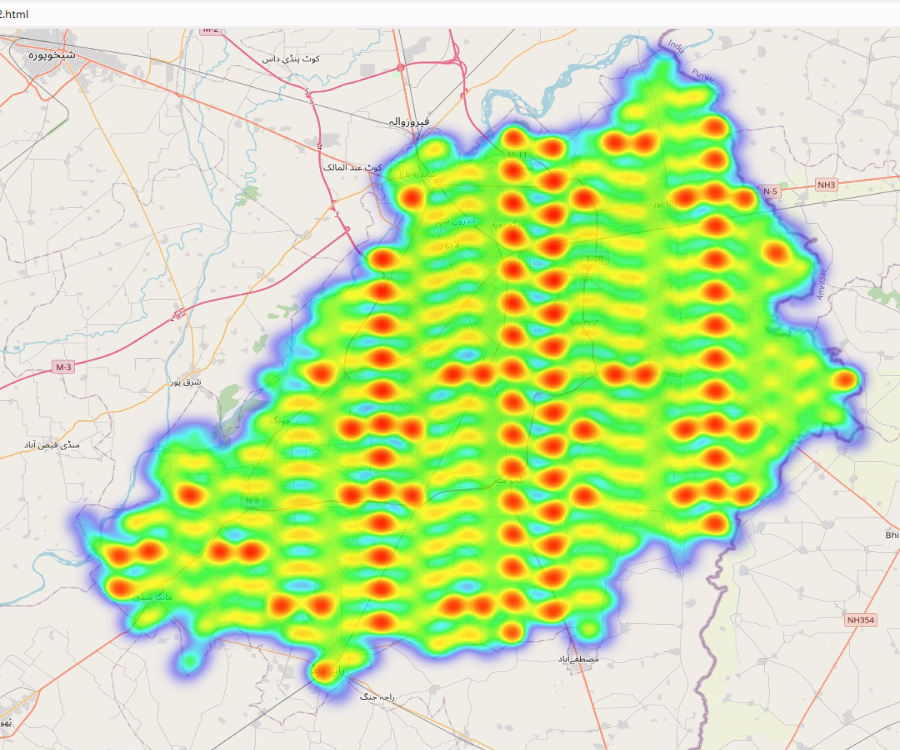}{Figures/PNG/frame_st_shaded_}{1}{9}
    \caption{AQI spatiotemporal map for the Lahore region. For optimal visualization, we recommend using Adobe Reader.}
    \vspace{-2mm}
    \label{fig:high-res-map}
\end{figure}

\begin{figure}[!t]
 \begin{subfigure}[b]{0.44\textwidth}
    \centering
    \includegraphics[width=1.0\linewidth]{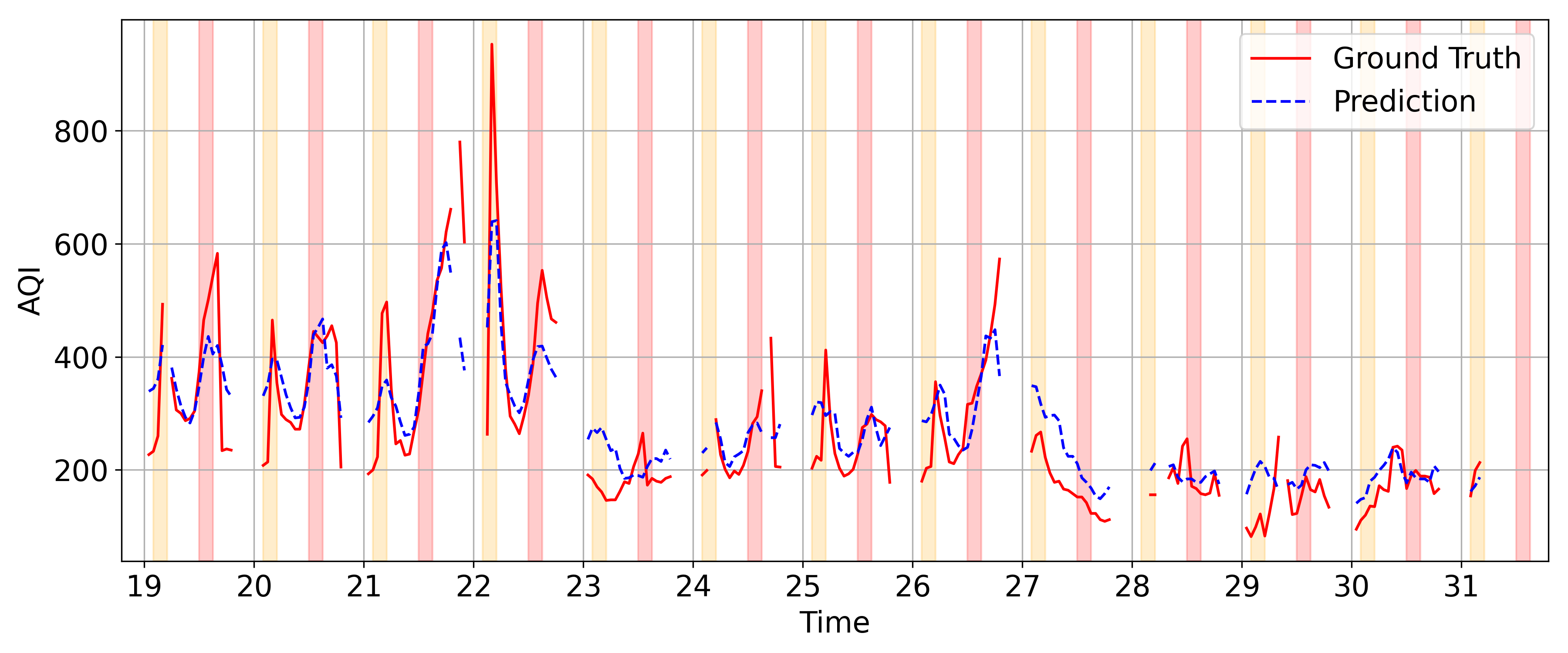}
    \caption{Sensor at node 657.}
    \label{fig:node_657}
    \end{subfigure}
   \begin{subfigure}[b]{0.45\textwidth}
    \centering
    \includegraphics[width=1.0\linewidth]{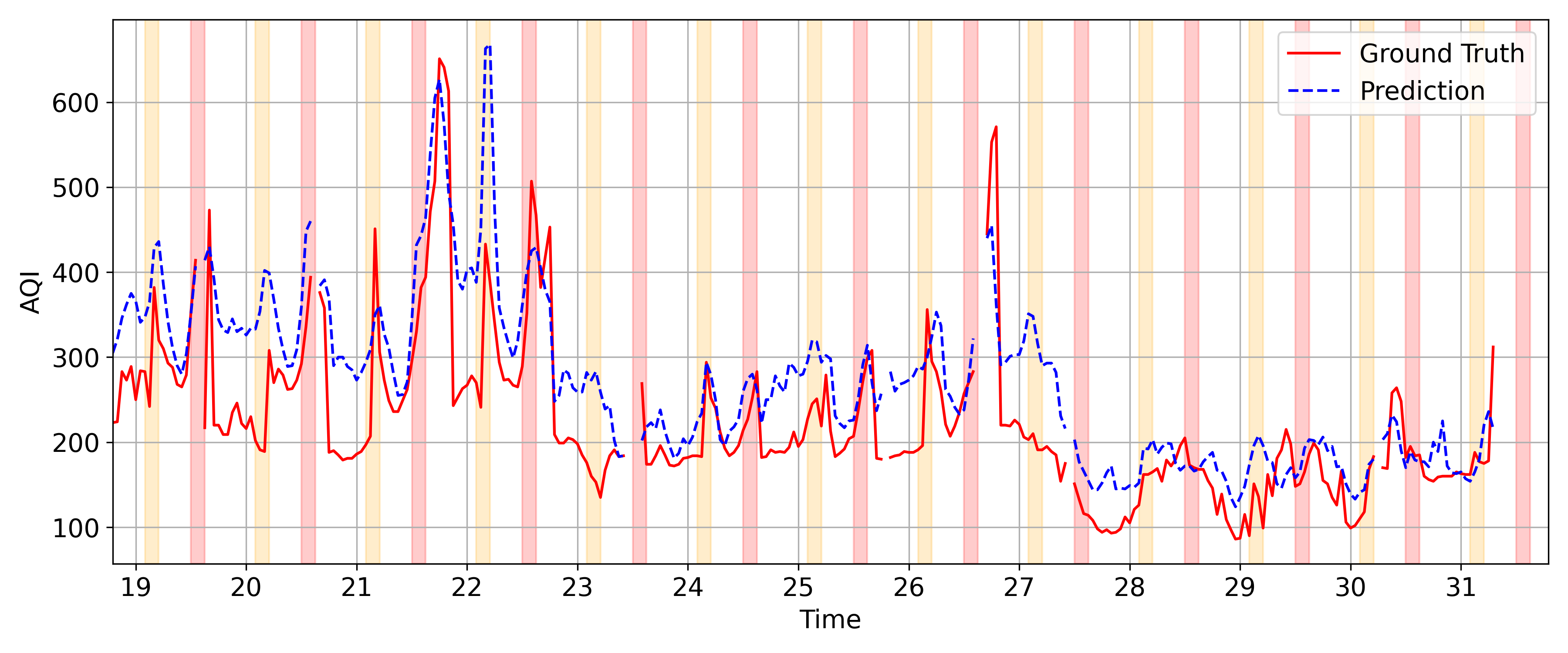}
    \caption{Sensor at node 1435.}
    \label{fig:node_1435}
    \end{subfigure}
    \caption{Comparison of ground truth and predicted AQI values for sensors at nodes (a) 657 and (b) 1435 from December 19 to 31, 2024, at an hourly resolution. The shaded regions represent rush hours: 7:00–10:00 (orange) and 17:00–20:00 (pink).}
    \vspace{-3mm}
    \label{fig:plots}
\end{figure}

Table~\ref{tab:parameter} presents performance metrics for two individual runs, as well as the average metrics computed over 10 runs. The results indicate that the best average performance is achieved at $lr = 10^{-4}$ and $\lambda = 10^{-6}$, underscoring the importance of both the learning rate and smoothness parameter in training the model effectively. The variability in performance metrics across runs can be attributed to differences in graph topology, as the connectivity and spatial dependencies within the graph significantly influence the learning process in graph-based models. Once the model is trained, we use it to generate high-resolution AQI spatial maps for the Lahore region. Figure~\ref{fig:high-res-map} illustrates this for various dates and times of the day. Notably, higher AQI values are observed on December 22 because of increased aerosol concentrations at $0.44~\mu m$ and $0.55~\mu m$, which indicates reduced sunlight exposure.
We compare the predicted AQI values with the actual AQI values for two sensor nodes during Run 2, as shown in Fig.\ref{fig:plots}. Morning rush hours (7:00–10:00) and evening peak hours (17:00–20:00) are highlighted to analyze real-time AQI patterns. Notably, AQI peaks during evening hours, driven by traffic congestion and reduced sunlight. While the model effectively captures real-time trends, it does not fully replicate the entire structure of the data. Specifically, the model under-predicts AQI during evening hours in Fig.\ref{fig:node_657} and overestimates it during morning rush hours in Fig.~\ref{fig:node_1435}.

The performance of the model deteriorates more significantly when the time points of $X_1$ are reduced from 4 to 2 days compared to reducing the time resolution of $X_3$ from 24 to 12 hours. This highlights the critical role of AOD and meteorological parameters at a daily resolution in estimating AQI values. Additionally, the performance of our deep neural network (DNN) improves with larger temporal dependencies, demonstrating that temporal data at multiple resolutions enhances the predictive capabilities of the model. While spatial features are crucial for capturing the dynamic characteristics of urban environments, certain features limits the generalizability of the model. For example, Points of Interest (PoIs) and Urban Green Spaces (UGS) exhibit sparse distributions across the graph, which can negatively affect performance. To evaluate the contribution of individual spatial features, we trained the model with isolated feature sets. The analysis revealed that population counts have a more significant impact on AQI predictions compared to other spatial features. Despite the variability in their utility, spatial features remain integral to accurately modeling the complexities of large cities. Although we implemented this framework on a spatially sparse set of temporal AQI samples, it can be extended to handle long-duration datasets effectively.

\section{Conclusions}
We have proposed a framework for high-resolution spatiotemporal mapping of air quality using sparse sensor data, satellite imagery, and various environmental factors. By leveraging GNNs, our framework estimates AQI at unmonitored locations, capturing both spatial and temporal dependencies. We have integrated a wide range of features, including meteorological data, road networks, PoIs, population density, and urban green spaces, into the model. Our case study in Lahore demonstrated the effectiveness of the proposed approach of using multi-resolution data to generate AQI maps at a fine spatial scale. We believe that the proposed framework offers a valuable tool for policymakers to assess air quality distribution in densely populated areas and devise targeted interventions to mitigate pollution. Incorporating additional data sources and exploring different network architectures for improving the accuracy of AQI predictions are future research directions.


\small
\bibliographystyle{IEEEtranN}
\bibliography{references}

\end{document}